\documentclass[journal]{IEEEtran}

\ifCLASSOPTIONcompsoc
  \usepackage[nocompress]{cite}
\else
  \usepackage{cite}
\fi

\ifCLASSINFOpdf
\else
\fi

\usepackage{amsmath}

\usepackage{algorithmic}

\newcommand{\etal}{\textit{et al.}}
\usepackage{times}
\usepackage{epsfig}
\usepackage{graphicx}
\usepackage{amsmath}
\usepackage{amssymb}

\usepackage{enumitem}
\usepackage{tablefootnote}
\usepackage{multirow}
\usepackage{capt-of}
\usepackage{fixltx2e}
\usepackage{floatrow}
\usepackage{soul}
\usepackage{algorithmic}
\usepackage[ruled,vlined]{algorithm2e}
\graphicspath{{Figs1/}}

\begin{document}

\title{Learning to Restore a Single Face Image Degraded by Atmospheric Turbulence using CNNs}

\author{Rajeev Yasarla,~\IEEEmembership{Student Member,~IEEE} and 
	Vishal~M.~Patel,~\IEEEmembership{Senior Member,~IEEE}
	\thanks{Rajeev Yasarla is with the Whiting School of Engineering, Johns Hopkins University, 3400 North Charles Street, Baltimore, MD 21218-2608, e-mail: ryasarl1@jhu.edu}
	\thanks{Vishal M. Patel is with the Whiting School of Engineering, Johns Hopkins University, 3400 North Charles Street, Baltimore, MD 21218-2608, e-mail: vpatel36@jhu.edu}}

\markboth{Journal of \LaTeX\ Class Files,~Vol.~14, No.~8, August~2015}%
{Shell \MakeLowercase{\textit{et al.}}: Bare Demo of IEEEtran.cls for Computer Society Journals}

\IEEEtitleabstractindextext{%
\begin{abstract}
Atmospheric turbulence significantly affects imaging systems which use light that has propagated through long atmospheric paths.  Images captured under such condition suffer from a combination of geometric deformation and space varying blur.  We present a deep learning-based solution to the problem of restoring a turbulence-degraded face image where prior information regarding the amount of geometric distortion and blur at each location of the face image is first estimated using two separate networks.  The estimated prior information is then used by a network called, Turbulence Distortion Removal Network (TDRN), to correct geometric distortion and reduce blur in the face image.  Furthermore, a novel loss is proposed to train TDRN where first and second order image gradients are computed along with their confidence maps to mitigate the effect of turbulence degradation.  Comprehensive  experiments on synthetic and real face images show that this framework is capable of alleviating blur and geometric distortion caused by atmospheric turbulence, and significantly improves the visual quality.   In addition, an ablation study is performed to demonstrate the improvements obtained by different modules in the proposed method.
\end{abstract}

}

\maketitle

\IEEEdisplaynontitleabstractindextext

%
\IEEEpeerreviewmaketitle

\section{Introduction}\label{sec:introduction}
Atmospheric turbulence significantly affects the quality of long-distance imaging systems by causing spatially and temporally random fluctuations in the index of refraction of the atmosphere \cite{Turbulence_Book}.  Variations in the refractive index causes the captured image to be geometrically distorted and blurry.  As a result, the captured images are often of very poor quality.   These degradations in turn may affect the performance of many computer vision systems such as long-distance surveillance and autonomous vehicles. 

Under the assumption that the scene and the imaging sensor are both static and that observed motions are due to the air turbulence alone, the image degradation due to atmospheric turbulence can be mathematically formulated as follows \cite{furhad2016restoring}, \cite{zhu2012removing}, \cite{hirsch2010efficient}, \cite{Law_thesisPaper}
\begin{equation}	\label{eq:model}
T_{k} = D_{k}(H_{k}(I)) + \epsilon_{k}, \;\;\; k=1, \cdots, N,
\end{equation}		
where $I$ denotes the ideal face image, $T_{k}$ is the $k$-th observed frame, $D_{k}$ and $H_{k}$ represent the deformation operator and air turbulence-caused blurring operator, respectively, $\epsilon_{k}$ denotes additive noise and $N$ denotes the total number of observed frames.  The deformation operator is assumed to deform randomly and $H_{k}$ correspond to a space-invariant diffraction-limited point spread function (PSF).  As can be seen from \eqref{eq:model}, atmospheric turbulence has two main degradations on the observed face images:  geometric distortion and blur.   Various optics-based \cite{pearson1976atmospheric,tyson2015principles,Turbulence_Book} and image processing-based \cite{metari2007new,shimizu2008super,furhad2016restoring,meinhardt2014implementation,micheli2014linear,zhu2012removing,lau2019restoration,chak2018subsampled} turbulence removal algorithms have been proposed in the literature.  In particular, many previous approaches have specifically focused on restoring an image from a given image sequence or video distorted by atmospheric turbulence \cite{zhu2012removing,lau2019restoration,chak2018subsampled}.  These methods make use of approaches such as  sub-sampling with Beltrami coefficients or fusing the image sequence in order to obtain an intermediate sharp image.  These intermediate images are further processed to remove blur and obtain a sharp image.  Moreover these methods assume the scene or object in the image to be static with changing atmospheric turbulence flow, which may not be practical in real world applications as the objects appearing in the scene can have different motion with respect to the  camera.

\begin{figure*}[t!]
	\centering
	\includegraphics[width=0.11\textwidth]{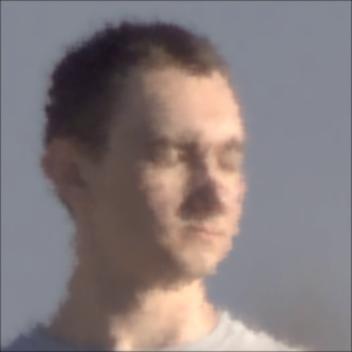}
	\includegraphics[width=0.11\textwidth]{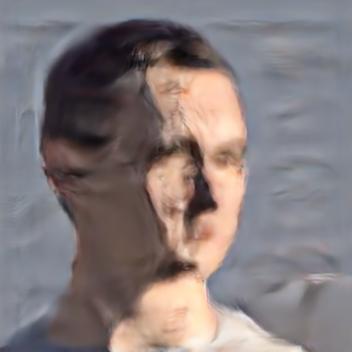} 
	\includegraphics[width=0.11\textwidth]{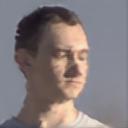}
	\includegraphics[width=0.11\textwidth]{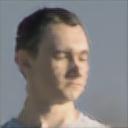}
	\includegraphics[width=0.11\textwidth]{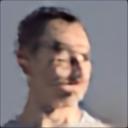}
	\includegraphics[width=0.11\textwidth]{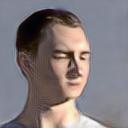}
	\includegraphics[width=0.11\textwidth]{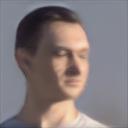} 
	\includegraphics[width=0.11\textwidth]{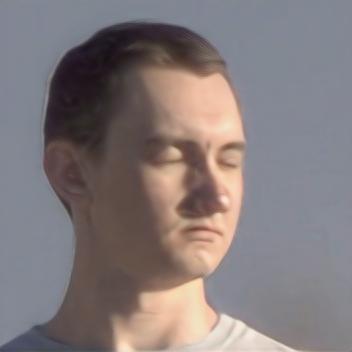}\\
	(a)\hskip50pt(b)\hskip50pt(c)\hskip50pt(d)\hskip45pt
	(e)\hskip50pt(f)\hskip50pt(g)\hskip50pt(h)\\
	\caption{Sample image restoration results. (a) Real turbulence distorted face image.  Results corresponding to (b) Pan et al.~\cite{pan2014deblurring}, (c) Shen et al.~\cite{shen2018deep},  (d) Yasarla et al.~\cite{yasarla2019deblurring}, (e) Pix2Pix~\cite{isola2017image}, (f) Kupyn et al.~\cite{kupyn2018deblurgan}, (g) Zhang et al.~\cite{zhang2019deep}, (h) Our proposed method, TDRN.}

	\label{Fig:exp1}
\end{figure*}

Since turbulence degradation is caused by a combination of geometric distortion and blur, it is difficult to restore a face image without any prior information.  Priors such as face exemplar masks, facial fiducial points and face semantic maps have been previously used in the literature to restore a blurry image \cite{pan2014deblurring,shen2018deep,yasarla2019deblurring}.  However, even when these methods are re-trained on turbulence distorted images they fail to recover sharper face images from degraded images.   This is mainly due to the fact that turbulence degradation involves a combination of both geometric distortion and blur and as a result these methods are not able to extract facial fiducial points or structural information like facial semantic maps reliably.  This can be clearly seen by comparing the performance of various recent image restoration algorithms on a real turbulence degraded face image as shown in Fig.~\ref{Fig:exp1}. Note that these image restoration algorithms are re-trained using turbulence-distorted images.  In this figure, the performance of methods that make use of priors as well as methods that do not make use of any priors are compared.  As can be seen from this figure, image restoration methods like Kupyn et al.~\cite{kupyn2018deblurgan}, and Zhang et al.~\cite{zhang2019deep} do not perform well due to the lack of prior information about the input face image. Their outputs are still blurry and distorted.  Similarly, methods that do make use of facial prior information do not perform well either \cite{pan2014deblurring,shen2018deep,yasarla2019deblurring}.   For instance, Pan et al.~\cite{pan2014deblurring} and Shen et al.~\cite{shen2018deep} produce the output images with artifacts on the face and  Yasarla et al.~\cite{yasarla2019deblurring} is not able to remove the distortions from the image completely.

\begin{figure*}[htp!]
	\begin{center}
		\centering
		\includegraphics[width=\textwidth]{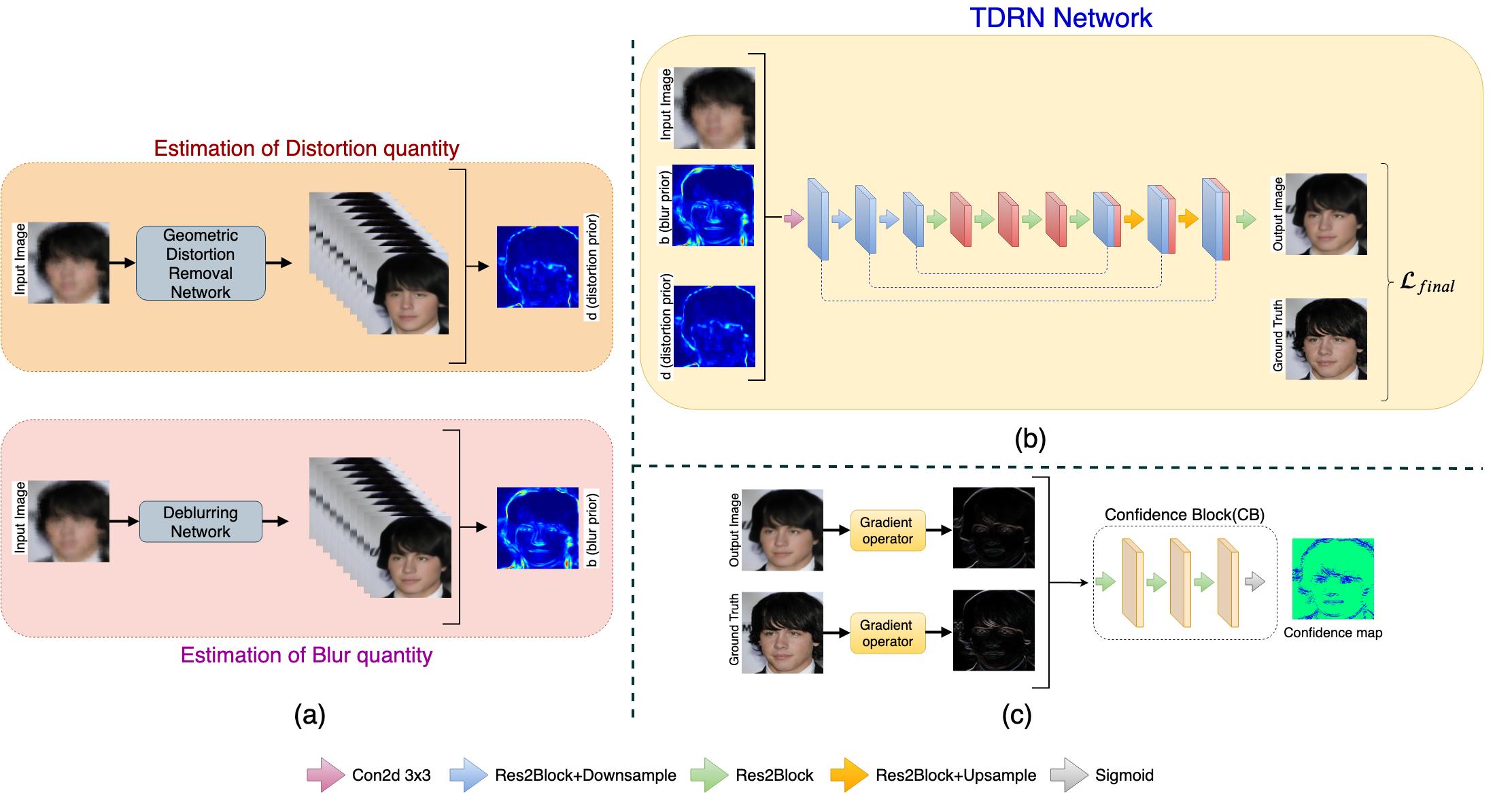}
		\caption{An overview of the proposed TDRN method. (a) Estimation of $b$ (blur prior) using the Deblurring Network (DBN) and $d$ (distortion prior) using the Geometric Distortion Removal Network (GDRN). (b)  The estimated $b$ and $d$ priors are used along with $T$ (turbulence distorted image) as an input to TDRN for obtaining a restored face image, $\hat{I}$. TDRN is trained with $\mathcal{L}_{final}$ which contains L2 loss, and $\mathcal{L}_{g}$ which is computed using $C_h,\: C_v,\: C_s$, obtained with the help of Confidence Blocks (CB) and gradient operators $\nabla_h$, $\nabla_v$ and $\nabla^2$. (c) An overview of  how confidence map is estimated using the Confidence Block (CB).}
		\label{Fig:TDRN}
	\end{center}
\end{figure*}

In many practical applications (i.e. surveillance), we may be faced with a scenario where we have to restore a single turbulence degraded image  (i.e $k=1$ in \eqref{eq:model}).   In this case, the lack of temporal information needed to minimize the variation in distortion caused by turbulence makes the problem more challenging. In this paper, we address the more practical problem of restoring a single face image distorted by turbulence and provide a deep learning-based solution. In particular, we propose a novel network called Turbulence Distortion Removal Netwrok (TDRN). Figure~\ref{Fig:TDRN} gives an overview of the proposed approach.   In our method, we estimate the prior information regarding blur and geometric distortion using two separate networks that are specifically trained for the individual tasks like deblurring and removal of geometric distortion.  Using Monte Carlo dropout in the deblurring network that is trained only for removing blur, we formulate epistemic uncertainty (defined in Kendall et al.~\cite{kendall2017uncertainties}), and use it as a prior information that resembles the amount of blur at each pixel in the image. Similarly, using Monte Carlo dropout in geometric distortion removal network, we compute a prior that resembles the amount of distortion at each location in the image. These priors along with the distorted image are used as input to TDRN to obtain a restored face image. 

It is commonly acknowledged in~\cite{NIPS2014_artifacts} that CNN-based methods trained using only the L2 loss often suffer from over-smoothing, ringing and jagged artifacts (which we refer to as halo artifacts).  To address this issue, we propose to use a new edge-preserving loss that makes use of  the first and second order image gradients.   Furthermore,  confidence scores at each pixel of the first and second order gradient images are calculated and used to re-weight the edge-preserving loss function.  Note that these confidence scores indicate how confident the TDRN network is about its computation at each pixel. Also these confidence scores help the TDRN network in learning the weights to remove the turbulence distortions.

Extensive experiments are conducted on two synthetic datasets (Helen~\cite{liu2015deep} and CelebA \cite{le2012interactive}) and one real-world face dataset.  In addition, comparisons are performed against several recent state-of-the-art image restoration approaches and it is demonstrated that TDRN is capable of alleviating blur and geometric distortions caused by turbulence, recovering details of the face with improved  visual quality.  Furthermore, an ablation study is conducted to demonstrate the improvements obtained by different modules in the proposed network.  Figure~\ref{Fig:exp1}(h) presents a sample output from our network, where one can clearly see that TDRN recovers the details of the face and significantly improves the visual quality better than the other image restoration methods.    To the best of our knowledge, this is one of the first deep learning-based methods to address the removal of atmospheric turbulence from a single face image. 
In summary, this paper makes the following contributions:
\begin{itemize}[noitemsep]
	\item We propose a novel way of addressing the removal of turbulence distortions using the priors that resemble the amount of blur and geometric distortions at each pixel.
	\item We introduce a way to compute blur and distortion priors using the networks that are trained for the individual tasks like deblurring and geometric distortion removal separately.
	\item Extensive experiments are conducted on two synthetic datasets and one real-world face image dataset.  In addition, comparisons are performed against several recent state-of-the-art image restoration approaches. Furthermore, an ablation study is conducted to demonstrate the improvements obtained by different modules in TDRN.
\end{itemize}

\section{Related Work}


\noindent {\bf{Video-based methods:}}  Since the information in a single frame is usually insufficient to restore an image, most approaches in the literature are based on videos or a sequence of images under the assumption that the scene is static.  Methods such as \cite{aubailly2009automated,vorontsov2001anisoplanatic} follow  a ``lucky frame" approach where a single frame is selected or multiple frames are fused to restore the image from turbulence degraded observations.  
Anantrasirichai et al. \cite{anantrasirichai2013atmospheric} address the problem by extracting the regions of interest from a few good frames and fuse them using the dual tree complex wavelet transform.  A few other approaches in the literature \cite{hirsch2010efficient,lou2013video,zhu2012removing,lau2019restoration,xie2016removing} attempt to tackle the deturbulence problem using registration-fusion approach where one computes a good reference image and then aligns the distorted frames in the video using a non-rigid registration algorithm. Lou et al.~\cite{lou2013video} sharpen each frame using the spatial Sobolev gradient flow and use temporally smoothing to minimize deformation between frames of a video. Zhu et al.~\cite{zhu2012removing} apply  a B-spline nonrigid registration algorithm and a patch-wise temporal kernel regression to obtain a turbulence free image. Lau et al.~\cite{lau2019restoration} propose a Robust Principle Component Analysis (RPCA) based approach on the deformation fields between the image frames and warp the image frames by a quasiconformal map associated with the low-rank part of the deformation matrix to obtain a sharp image. Xie et al.~\cite{xie2016removing} propose a hybrid method which assigns the low-rank image to be the initial reference image and then a variational model is used to improve the quality of the reference image.  Finally, the frames are registered using the obtained reference image. Recently, Chak et al.~\cite{chak2018subsampled} proposed a deep learning-based approach where a sub-sampling algorithm is incorporated into the deep network to filter out strongly corrupted frames to obtain an improved restored image.\\


\noindent {\bf{Image deblurring methods: }}	Since the images captured under atmorspheric turbulence suffer from both blur and geometric deformation, we briefly review some recent deep learning-based image deblurring methods.  We limit our discussion to image deblurring methods that are specifically designed for restoring blurry face images which use some prior information about the face.    Pan et al.~\cite{pan2014deblurring}   extract  the edges of face parts and estimate exemplar face images, which are further used as the global prior to estimate the blur kernel for image restoration.  Shen et al.~\cite{shen2018deep} exploit global semantic priors and local structural constraints to perform face deblurring using a multi-scale CNN and adversarial training \cite{radford2015unsupervised}. Lu et al.~\cite{lu2019unsupervised} proposed a domain-specific single face image deblurring method by disentangling the content information in an unsupervised fashion using the KL-divergence.  Very recently Yasarla \etal \cite{yasarla2019deblurring} proposed  a multi-stream  architecture  and training  methodology  that  exploits  facial semantic  labels  for  image   deblurring.  Recently, Lau \etal proposed a turbulence removal method which disentangles   the   blur   and   geometric deformation   due   to   turbulence  and  reconstructs  a  restored  image.  This is done by decomposing  geometric  deformation and blur effects using two seperate generators.  Most image deblurring networks are based on either UNet \cite{zhang2019deep,yasarla2019deblurring} or ResNet \cite{kupyn2018deblurgan,shen2018deep} architectures.  As a result, though they are specifically designed for image deblurring, one can re-train them on the turbulence degraded data to restore a single image degraded by atmospheric turbulence.   

\section{Proposed Method}
The mathematical model presented in \eqref{eq:model}, is for restoring  a sharp image from a sequence of turbulence-degraded images.  In this paper, we assume the availability of only a single distorted frame (i.e $k=1$) to reconstruct the clean face image.  In this case, we have the following observation model 
\vskip-10pt
\begin{equation}	\label{eq:general_model}
T = D(H(I)) + \epsilon,
\end{equation}	
where the subscript $k$ has been removed from \eqref{eq:model}.  This is an extremely ill-posed problem as we have to overcome the effects of both blur and geometric distortion from a single image.   Some prior information is needed to recover $I$ from $T$.  As discussed earlier, state-of-the-art face image restoration methods that use some prior information about the face do not perform well on turbulence degraded images even though they are retrained on the turbulence degraded images (see Fig.~\ref{Fig:exp1}).  This is mainly due to the fact that in turbulence degraded images, the facial componenets and texture information are severly degraded.  As a result, it is difficult to extract facial semantic maps or exemplar masks from these turbulence degraded images. This lead to poor performance of face deblurring methods like \cite{pan2014deblurring,shen2018deep,yasarla2019deblurring} which extract facial information in the form exemplar masks or semantic information, and use them as priors.  In order to address this problem, we propose a way to estimate the amount of blur and geometric distortion that each pixel undergoes in the degraded face image and use them as priors to recover the clean image. In what follows, we first describe how these priors are estimated.

\begin{figure*}[htp!]
	\begin{center}
		\centering
		\includegraphics[width=\textwidth]{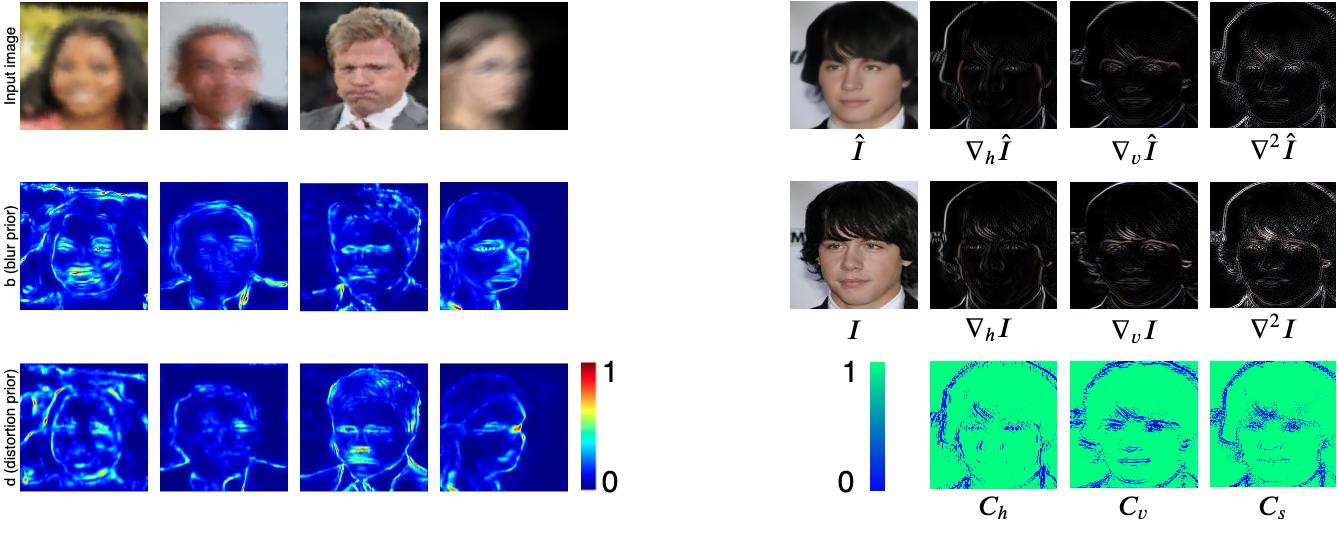}\\
		(a)\hskip225pt (b)
		\caption{(a) First row: sample turbulence distorted images. Second row: corresponding blur priors $b$'s. Third row:  corresponding distortion priors $d$'s. (b) Confidence maps corresponding to the first and second order image gradients.}
		\label{Fig:priors}
	\end{center}
	
\end{figure*}

\subsection{Blur and Geometric Distortion Prior Estimation}\label{priors_est}
We compute a blur prior, $b$ and a distortion prior, $d$ using two separate networks that are specifically trained for image deblurring and geometric distortion removal, respectively (see Figure~\ref{Fig:TDRN}(a)). The image deblurring network, DBN, and the geometric distortion removal network, GDRN, are constructed using a combination of UNet~\cite{ronneberger2015u} and DenseNet~\cite{huang2017densely} architectures with Res2Block as the basic building block \cite{gao2019res2net}.  DBN and GDRN are similar networks which consist of the following sequence of layers,\\
Res2Block(3,64)-Downsample-Res2Block(64,64)-Downsample-ResBlock(64,64)\\
-Res2Block(64,64)-Res2Block(64,64)-Res2Block(64,64)-Res2Block(64,64)\\
-Upsample-Res2Block(64,64)-Upsample-Res2Block(64,16)-Res2Block(16,3),\\
where Downsample means average pooling layer and  Res2Block$(m,n)$ means Res2Block with $m$ input channels and $n$ output channels. 

Let $DBN(.;\theta_{db})$ and $GDRN(.;\theta_{gd})$ denote the DBN and GDRN networks with parameters $\theta_{db}$ and $\theta_{gd}$, respectively.  By applying Monte Carlo dropout in every layer of the DBN and GDRN, we can formulate the epistemic uncertainty \cite{kendall2017uncertainties} and use the corresponding variance as a prior information.  For example, given a turbulence distorted image $T$, we pass it as an input to DBN $S$ times and obtain a set of outputs $\{P_i\}_{i=1}^S$, where $i$ corresponds to the $i$th instance of dropout and $P_i = DBN(T;\theta^{i}_{db})$. We define the blur prior, $b$ as the variance of the outputs $\{P_i\}_{i=1}^S$, i.e. $b = variance(\{P_i\}_{i=1}^S)$. As explained by Kendall et al.~\cite{kendall2017uncertainties} this variance is defined as the model uncertainty. However, in our case we use it as a measure of the ability or competence of  the network in addressing image deblurring. Hence a high variance value at a pixel location in $b$ means that DBN is not able to reconstruct the underlying clean image properly at that pixel location in the output image.  Similarly, we pass $T$ as an input to GDRN $S$ times and obtain a set of outputs $\{Q_i\}_{i=1}^S$, where $i$ corresponds to the $i$th instance of dropout and $Q_i = GDRN(T;\theta_{gd}^{i})$.  We define the distortion prior, $d$ as the variance of the obtained outputs, i.e. $d = variance(\{Q_i\}_{i=1}^S)$.   Fig.~\ref{Fig:priors}(a) shows some sample turbulence distorted images and their corresponding priors $b's$ from DBN and the distortion priors $d's$ from GRDN.

\subsection{Turbulence Distortion Removal Network (TDRN)}\label{network}

As discussed earlier, given a turbulence distorted image, $T$, priors $b$ and $d$ are computed using DBN and GDRN, respectively and are passed as inputs to TDRN along with $T$ to obtain the restored image, $\hat{I}$ (see Figure~\ref{Fig:TDRN}(b)).  Let $TN(.)$ denote the TDRN network.  As a result, $\hat{I}= TN(T,b,d)$.

The turbulence distortion removal network  is constructed using the  UNet~\cite{ronneberger2015u}  architecture with Res2Block as the basic building block \cite{gao2019res2net}. TDRN consists of the following sequence of convolutional layers,\\
Conv2d $3\times 3$(5,16)- Res2Block(16,64)- Downsample-Res2Block(64,64)\\
-Downsample-Res2Block(64,64)-Res2Block(64,64)-Res2Block(64,64)\\
-Res2Block(64,64)-Res2Block(64,64)-Upsample-Res2Block(64,64)\\-Upsample-Res2Block(64,3),\\
where Conv2d $3\times 3(m,n)$ is a $3\times 3$ convolutional layer with $m$ input channels and $n$ output channels.

\subsubsection{Loss to train TDRN}
To make sure that the restored image preserves sharp edges and avoids halo artifacts, an edge-preserving loss function with confidence is proposed in this paper based on the first and second order image gradients.  In particular, we make use of the aleotroic uncertainty formulation  \cite{kendall2017uncertainties} which is data dependent and defines pixel wise confidence values which indicate how confident the network is about its computation to calculate the confidence maps.  Given the ground truth image, $I$ and the restored image $\hat{I}$ we compute the first order image gradients in the vertical direction $(\nabla_v{I}, \nabla_v\hat{I})$ and pass them to a Confidence Block (CB) as shown in Figure~\ref{Fig:TDRN} to obtain the corresponding confidence map $C_v$.  Similarly, we calculate the image gradients in the  horizontal direction $(\nabla_h{I}, \nabla_h\hat{I})$ as well as the second order gradients using the Laplace operator ($\nabla^2{I}, \nabla^2\hat{I}$) and pass them to CB to obtain the corresponding confidence maps, $C_{h}$ and $C_{s}$, respectively.  CB consists of a sequence of three Res2Blocks followed by a Sigmoid layer.  Confidence maps corresponding to the first and the second order gradient images are shown in Figure~\ref{Fig:priors}(b), where it can be clearly seen that the confidence values are low for those pixels that have high errors in the corresponding gradient images.  Low values in a confidence map indicate high reconstruction error in the corresponding gradient image.  The proposed confidence-based  edge-preserving loss guides the TDRN network to learn appropriate weights so that the reconstruction error is minimized in the high reconstruction error region of the gradient image.  Note that the confidence values in the confidence map take values between 0 and 1. The confidence-based loss on the first and second order gradients is defined as follows

\begin{multline}
\mathcal{L}_g = 
\sum_{i=h,v,s} \sum_{p=1}^{W_1}\sum_{q=1}^{W_2} ( C_i(p,q)\lVert \nabla_i \hat{I}(p,q) - \nabla_i I(p,q) \rVert_1\\  - \lambda_c \log C_i(p,q) ),
\end{multline}
where $\lambda_c$ is a constant and we assume that the images are of size $W_{1}\times W_{2}$.  We set  $\lambda_c$ equal to  $0.01$ while training TDRN. This formulation of the loss function is motivated by the loss proposed by Kendall  et al.~\cite{kendall2017uncertainties}.

In addition to $\mathcal{L}_g$ we use the  perceptual  loss to train our network.  Let $F(.)$ denote the features obtained using the VGG-Face model \cite{parkhi2015deep}, then the perceptual loss is defined as follows, $\mathcal{L}_p = \| F(\hat{I})-F(I)\|_2^{2}.$  Features from layer $pool5$ of a pretrained VGG-Face network \cite{parkhi2015deep} are used to compute the perceptual loss.  Finally, the overall loss used to train TDRN is a combination of the confidence guided loss, perceptual loss and L1-loss as follows,

\begin{equation}
\mathcal{L}_{final} = \mathcal{L}_1 + \lambda_g \mathcal{L}_g + \lambda_p \mathcal{L}_p,
\end{equation}
where $\lambda_g$ and $\lambda_p$ are constants and $\mathcal{L}_1$ is the L1 loss between the restored image and the ground truth clean image defined as follows, $\mathcal{L}_1 = \|\hat{I} - I\|_1 = \|TN(T,b,d) - I\|_1,$
where $\hat{I}$ is the restored image  using TDRN (i.e. $\hat{I} = TN(T,b,d)$).

\section{Experimental Results}
In this section, we present the experimental details and evaluation results on both synthetic and real-world datasets.  The performance of different methods on the synthetic data is evaluated in terms of Peak Signal-to-Noise Ratio (PSNR), Structural Similarity Index (SSIM) \cite{wang2004image} and $d_{VGG}$.  Here, $d_{VGG}$ corresponds to feature distance between the restored image and the ground truth clean image.  We use the outputs of $pool5$ layer from the VGG-Face~\cite{parkhi2015deep} to compute $d_{VGG}$. 
Furthermore, in order to show the significance of different face restoration methods, we perform face recognition on the restored images using ArcFace \cite{deng2019arcface}.    The proposed TDRN method is compared with the following recent state-of-the-art face image restoration methods \cite{pan2014deblurring,shen2018deep,yasarla2019deblurring} and generic single image deblurring methods \cite{kupyn2018deblurgan,zhang2019deep}.  Note that we re-train these methods using the same turbulence degraded images that are used to train our network.  While re-training the networks we followed the training procedure including the parameter selection mentioned in those respective papers.


\subsection{Training and Testing Datasets}
\label{Train}
\noindent\textbf{Training Datasets}: Based on the observation model \eqref{eq:general_model}, the authors of \cite{chak2018subsampled} have presented an efficient way of generating turbulence distorted images.  This procedure includes a variety of different parameters that control the distortion level. In our experiments, we use this approach to generate synthetic data using $128 \times 128$ cropped clean face images from the CelebA \cite{le2012interactive} and Helen \cite{liu2015deep} datasets.  We use 50,000 face images from the CelebA dataset train set  and 2,000 images from the Helen dataset train set to generate synthetic turbulence distorted images. 

Regarding the blur operator $H$, we generate Gaussian and motion blur kernels, and use them to blur the clean images. We use 8 isotropic kernels and 8 anisotropic Gaussian blur kernels from~\cite{zhang2018learning,zhang2019deep} with standard deviations uniformly sampled in the interval $[1,4]$ to generate blurry images. We follow the procedure described in DeblurGAN~\cite{kupyn2018deblurgan} to generate 30,000 motion blur kernels with blur kernel size varying from [11,27], and use them to obtain motion blurred face images.  Regarding the deformation operator, $D$, we select the following hyper-parameters \cite{chak2018subsampled}: standard deviation $\sigma=16$, distortion strength value $\eta =0.15$, number of iterations $M=\{1000,4000,7000,10000,13000,16000,19000\}$, and patch size $N=4$.  Given a clean face image $I$, we first blur it using the operator $H$ using the above mentioned blur kernels and then distort the face image using the operator $D$.   Finally we add Gaussian noise $\epsilon$ with standard deviation 0.02 to obtain the turbulence distorted image. In total, we obtain 1.5 million pairs of turbulence distorted and corresponding clean images $\{T_{i},I_{i}\}_{i=1}^{1.5\times10^6}$ to train our TDRN network. Note that, we denote these training set by $\mathcal{D}_{train}$ (i.e $\mathcal{D}_{train}$ contains 1.5 million pairs of $\{T,I\}$  images).\\

\noindent\textbf{Synthetic Datasets}: We create two different  synthetic testing datasets using the test images provided in the Helen \cite{liu2015deep} and CelebA  \cite{le2012interactive} datasets. We randomly pick 100 identities from test images from each of these datasets and generate the distorted images. We follow the procedure in DeblurGAN~\cite{kupyn2018deblurgan} to generate 16,000 motion blur kernels uniformly sampling the blur kernels in the interval~[13,27]. We use 8 isotropic kernels and 8 anisotropic Gaussian blur kernels from~\cite{zhang2018learning,zhang2019deep} with standard deviations uniformly sampled from the interval $[1,4]$ to obtain 8000 blurry images. Using these motion and Gaussian blur kernels, we blur the clean face images. These blurry images are then distorted using the operator $D$, with hyper-parameters $\sigma=16$, $\eta=0.15$, $N=4$ and $M=\{1000,6000,11000,16000,21000\}$. As a result, we generate 24,000 images from Helen and 24,000 images from CelebA  as test images (i.e. in total 48,000 test images denoted by $\mathcal{D}_{test}$) . 

\begin{table}[htp!]
	\caption{Details of atmospheric properties while capturing the real-world turbulence distorted images.
		\label{Table:real_tur}}
	\centering
	\resizebox{\textwidth}{!}{
		\begin{tabular}{|c|c|c|c|c|c|c|}
			\hline
			\begin{tabular}[c]{@{}c@{}}Atmospheric \\ Properties\end{tabular} &
			\begin{tabular}[c]{@{}c@{}}Distance \\from camera \\ (meters)\end{tabular} &
			\begin{tabular}[c]{@{}c@{}}Temperature\\ ($^\circ$C)\end{tabular} & \begin{tabular}[c]{@{}c@{}}Humidity\\ (\%)\end{tabular} & \begin{tabular}[c]{@{}c@{}}Wind speed\\ (meters/second)\end{tabular} & \begin{tabular}[c]{@{}c@{}}Wind direction\\ (radians)\end{tabular} & \begin{tabular}[c]{@{}c@{}}Atmospheric \\ Pressure \\ (millibar)\end{tabular}  \\ \hline 
			Value range  &  300-600 & 16.5-38.8   & 5-28.5   & 0.57-9.34  & 0-2$\pi$  & 919.3-931.2   \\ \hline 
	\end{tabular}}
\end{table}

\begin{figure}[htp!]
	\centering
	\includegraphics[width=\textwidth]{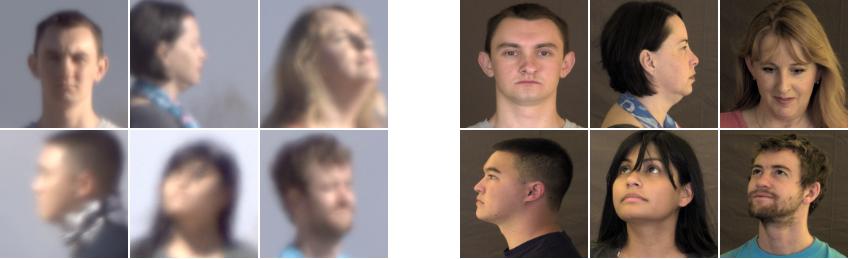}\\
	(a)\hskip120pt(b)
	\vskip -10pt
	\caption{(a) Sample real-world atmospheric turbulence distorted images. (b) Clean images from the gallery set.}
	\label{Fig:sample}
\end{figure}

\noindent\textbf{Real-world Dataset}: We use a test set consisting of 600 real-world turbulence distorted images collected by the US Army in a variety of different atmospheric conditions.   Table~\ref{Table:real_tur} shows the details of the atmospheric properties while the data was captured. Note that these 600 images correspond to 100 different individuals in 6 different poses.  The images were collected at the distances of 300 and 600 meters and the camera had a random motion while capturing the images.  As we don't have the corresponding ground truth images, we use face recognition to evaluate the performance of different methods.  Another set of 600 clean images of the same 100 subjects were also captured in a closed room. These images serve as the gallery set for face recognition.   We use Arcface  \cite{deng2019arcface} to perform face recognition.  The faces are cropped and aligned using the method proposed in~\cite{zhang2016joint}. Fig.~\ref{Fig:sample} shows some sample real-world turbulence distorted images and gallery images from this dataset.

\begin{algorithm}
	\label{alg:train}
	\SetAlgoLined
	\KwIn{Atmospheric turbulence distorted and corresponding clean image pairs $\{T_{j},I_{j}\}\in\mathcal{D}_{train}$, $GDRN(.;\theta_{gd})$, $DBN(.;\theta_{db})$ and initial TDRN network parameters $\tilde{\theta}_{td}$.}
	\KwResult{${\theta}_{td}$, optimized network parameters of TDRN}
	\For{every epoch }{
		\For{$\{T_{j},I_{j}\}\in\mathcal{D}_{train}$}{
			$b_{j}$ = variance($\{DBN(T_{j};\theta^{i}_{db})\}_{i=1}^{S}$);\\ $d_{j}$ = variance($\{GDRN(T_{j};\theta^{i}_{gd})\}_{i=1}^{S}$)\;
			$\hat{I}_{j}= TN(T_{k},b_{j},d_{j})$; update ${\theta}_{td}$ $\leftarrow$ using $\mathcal{L}_{final}$
			
		}
	}
	\caption{Pseudo code for training TDRN.}
\end{algorithm}

\subsection{Training and Testing Details} 
\noindent {\bf{TDRN Training:}}
Algorithm~\ref{alg:train} and Algorithm~\ref{alg:test} illustrate the procedures followed for training and testing TDRN, respectively.   Given $T_{k}$, we estimate the blur prior $b_{k}$ and the distortion prior $d_{k}$ using DBN and GDRN, respectively. Note that we set $S = 10$ while estimating the priors.  $\{T_{k},b_{k},d_{k}\}$ are passed as an input to TDRN to obtain $\hat{I}_{k}$.  TDRN is trained using $\mathcal{L}_{final}$. 	Note that, parameters of DBN and GDRN are kept frozen  while training TDRN.  In other words DBN and GDRN are used only to compute $b_{k}$ and $d_{k}$, respectively. We set $\lambda_c=0.01$, $\lambda_g=0.25$, and $\lambda_p=0.1$. TDRN is trained using the Adam optimizer with learning rate of 0.0002 and a batchsize of 10. TDRN is trained for $1.5 \times 10^5$ iterations.

\begin{algorithm}
	\label{alg:test}
	\SetAlgoLined
	\KwIn{Turbulence distorted images $T_{j}\in\mathcal{D}_{test}$,  $GDRN(.;\theta_{gd})$, $DBN(.;\theta_{db})$, TDRN $TN(.,.,.;{\theta}_{td})$.}
	\KwOut{Restored face image $\hat{I}_{j}$}
	\For{$T_{j}\in \mathcal{D}_{test}$}{
		$b_{j}$ = variance($\{DBN(T_{j};\theta^{i}_{db})\}_{i=1}^{S}$);\\
		$d_{j}$ = variance($\{GDRN(T_{j};\theta^{i}_{gd})\}_{i=1}^{S}$)\;
		$\hat{I}_{j}= TN(T_{j},b_{j},d_{j})$
	}
	\caption{Pseudo code for testing TDRN.}
\end{algorithm}

\noindent {\bf{DBN Training:}} DBN is trained using $\{I,H(I)\}$, where $I$ is the clean face image and $H(I)$ is the corresponding blurry image.  The blurry images are generated using the Helen and CelebA datasets as explained earlier.  The blur kernels used for $H$ are the same kernels as explained in   subsection~\ref{Train}.

\noindent {\bf{GDRN Training:}} GDRN is trained using $\{I,D(I)\}$, where $I$ is the clean face image and $D(I)$ is the corresponding geometrically distorted image.  The geometric distorted images are generated using the Helen and CelebA datasets. The hyper-parameters for the operator $D$ are set as follows:  $\sigma=16$, $\eta = 0.15$, $N=4$, and $M=\{1000,4000,7000,10000,13000,16000,19000\}$. 
The L1-loss is used to train both DBN and GDRN networks networks.  Adam optimizer with learning rate of 0.0002 and a batchsize of 10 is used during training. Both networks are trained for $1 \times 10^5$ iterations.

\begin{table}[hp!]
	\caption{Quantitative results in terms of PSNR, SSIM, and $d_{VGG}$ on the synthetic datasets. PSNR/SSIM higher the better, and $d_{VGG}$ lower the better} \label{Table:CompSt}
	\centering 
	\resizebox{\textwidth}{!}{
		\begin{tabular}{|l|c|c|c|c|c|c|}
			\hline
			\multirow{2}{*}{\begin{tabular}[c]{@{}c@{}}Deturbulence\\ Method\end{tabular}} & \multicolumn{3}{c|}{CelebA}   & \multicolumn{3}{c|}{Helen}   \\ \cline{2-7} 
			& PSNR   & SSIM    & $d_{VGG}$   & PSNR   & SSIM    & $d_{VGG}$   \\ \hline
			Turbulence-distorted   &  22.43    &    0.731   &   5.13      &   22.35     &   0.667     &  6.11       \\ \hline
			\multicolumn{1}{|l|}{Pix2Pix~\cite{isola2017image}(CVPR 2017)}   &  22.51 &  0.738  & 5.28 & 22.62    &  0.671    & 5.83 \\ \hline
			Pan et al.~\cite{pan2014deblurring}(ECCV 2014)    &   20.73    &   0.679  &  6.04    &    20.01    &   0.627    &  7.28  \\ \hline
			Shen et al.~\cite{shen2018deep}(CVPR 2018)   &   23.08      &    0.745 &  4.72   &  23.01  &   0.681 &  5.14 \\ \hline
			Yasarla et al.~\cite{yasarla2019deblurring}(TIP 2020)    & 24.06 &   0.768   &    4.11    &    23.81      &  0.702       &   4.49       \\ \hline
			Kupyn et al.~\cite{kupyn2018deblurgan}(CVPR 2018)      &   23.54     &   0.748    &   4.51   &   23.28   &   0.693     &  4.98   \\ \hline
			Zhang et al.~\cite{zhang2019deep}(CVPR 2019)     &  24.16    &   0.770 &  3.94     &   23.95    &    0.709    &   4.36     \\ \hline
			\multicolumn{1}{|l|}{TDRN (our method)}   &      \textbf{25.42} &    \textbf{0.815}  &  \textbf{3.09}    &  \textbf{25.08}   &   \textbf{0.752}  & \textbf{3.80} \\ \hline
	\end{tabular}}
\end{table}

\begin{figure*}[htp!]
	\centering
	\includegraphics[width=0.16\textwidth]{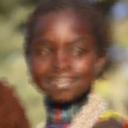}
	\includegraphics[width=0.16\textwidth]{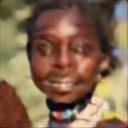} 
	\includegraphics[width=0.16\textwidth]{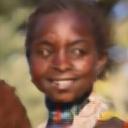}
	\includegraphics[width=0.16\textwidth]{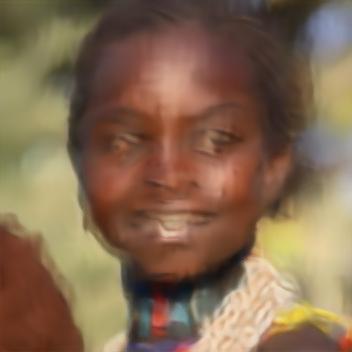}
	\includegraphics[width=0.16\textwidth]{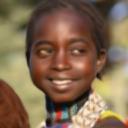}
	\includegraphics[width=0.16\textwidth]{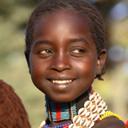}\\ \vskip3pt
	\includegraphics[width=0.16\textwidth]{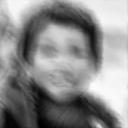}
	\includegraphics[width=0.16\textwidth]{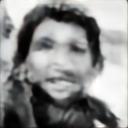} 
	\includegraphics[width=0.16\textwidth]{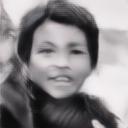}
	\includegraphics[width=0.16\textwidth]{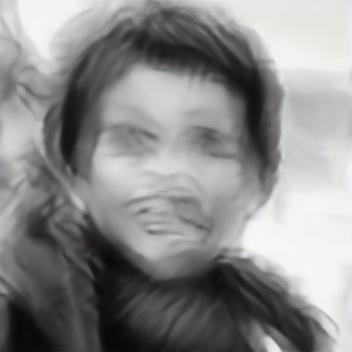}
	\includegraphics[width=0.16\textwidth]{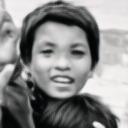}
	\includegraphics[width=0.16\textwidth]{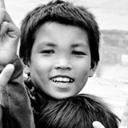}\\ \vskip3pt
	\includegraphics[width=0.16\textwidth]{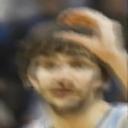}
	\includegraphics[width=0.16\textwidth]{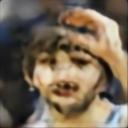} 
	\includegraphics[width=0.16\textwidth]{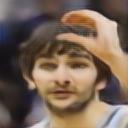}
	\includegraphics[width=0.16\textwidth]{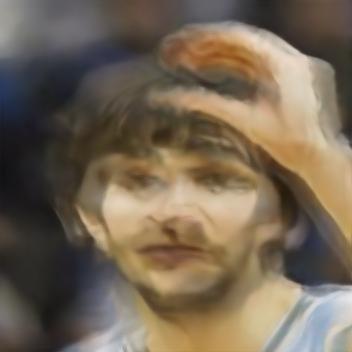}
	\includegraphics[width=0.16\textwidth]{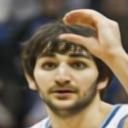}
	\includegraphics[width=0.16\textwidth]{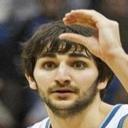}\\ \vskip3pt
	\includegraphics[width=0.16\textwidth]{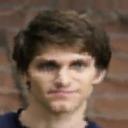}
	\includegraphics[width=0.16\textwidth]{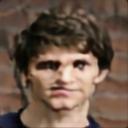} 
	\includegraphics[width=0.16\textwidth]{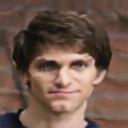}
	\includegraphics[width=0.16\textwidth]{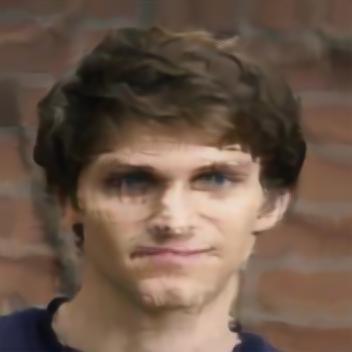}
	\includegraphics[width=0.16\textwidth]{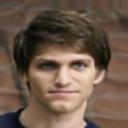}
	\includegraphics[width=0.16\textwidth]{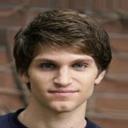}\\ \vskip3pt
	(a)\hskip75pt(b)\hskip75pt(c)\hskip75pt(d)\hskip75pt(e) \hskip75pt (f)
	\caption{Image restoration results on sample synthetic turbulence distorted images generated from the Helen dataset (first two rows) and the CelebA dataset (last two rows). (a) Turbulence distorted image.  results corresponding to (b) Pan et al.~\cite{pan2014deblurring}(ECCV 2014), (c) Shen et al.~\cite{shen2018deep}(CVPR 2018), (d) Zhang et al.~\cite{zhang2019deep}(CVPR 2019), (e) Our proposed method TDRN (f) ground-truth clean image. We can clearly observe that prior-based face deblurring methods Pan et al.~\cite{pan2014deblurring} and  Shen et al.~\cite{shen2018deep} produce artifacts on the restored face images. This is mainly due to the fact that their methods are not able to estimate the prior information like face semantic maps or face exemplar masks from the turbulence degraded images.  And using these improper priors while removing turbulence distortions results in artifacts. On the other hand Zhang et al.~\cite{zhang2019deep} which does not use any prior information produces blurry face images.  In comparsion with other approaches, the proposed TDRN method restores sharp and clean  face images.}
	\label{Fig:syn_CA}
\end{figure*}

\subsection{Results on Synthetic Datasets}
Results corresponding to different methods on synthetic datasets are shown in Figure~\ref{Fig:syn_CA} and Table~\ref{Table:CompSt}. The higher the PSNR/SSIM  and lower $d_{VGG}$, the better the quality of the  reconstructed image.  As can be seen from Table~\ref{Table:CompSt}, TDRN outperforms the state-of-the-art face image restoration methods.  In particular, the generic  deblurring methods~\cite{kupyn2018deblurgan,zhang2019deep} are not able to perform well due to the lack of prior information.  On the other hand,  methods that make use of some prior information about the face \cite{pan2014deblurring,shen2018deep,yasarla2019deblurring} are unable perform better because of improper prior estimation from the input images. The proposed method outperforms the state-of-the-art methods by about 1dB in PSNR, 0.04 in SSIM and 0.7 in $d_{VGG}$, which clearly demonstrates the effectiveness of the proposed method.


\begin{figure*}[htp!]
	\begin{center}
		\centering
		\includegraphics[width=0.135\textwidth]{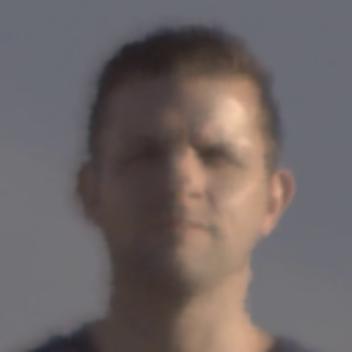}
		\includegraphics[width=0.135\textwidth]{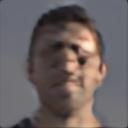} 
		\includegraphics[width=0.135\textwidth]{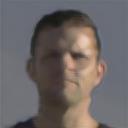}
		\includegraphics[width=0.135\textwidth]{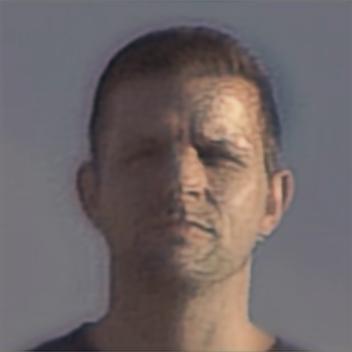}
		\includegraphics[width=0.135\textwidth]{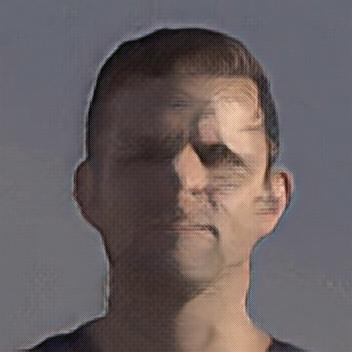}
		\includegraphics[width=0.135\textwidth]{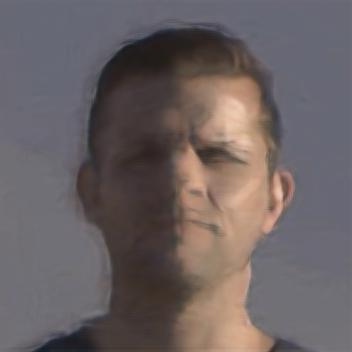}
		\includegraphics[width=0.135\textwidth]{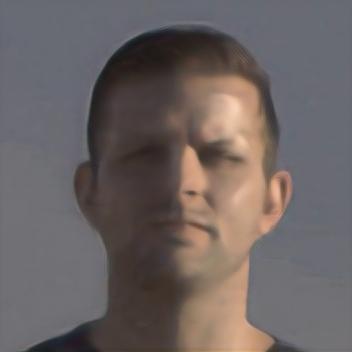}\\ \vskip3pt
		\includegraphics[width=0.135\textwidth]{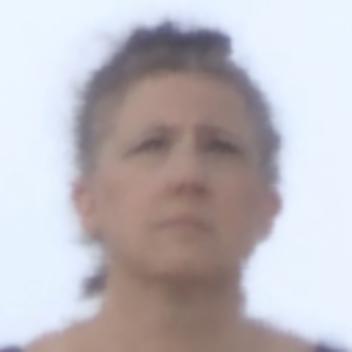}
		\includegraphics[width=0.135\textwidth]{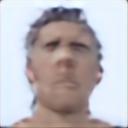} 
		\includegraphics[width=0.135\textwidth]{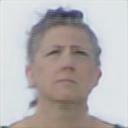}
		\includegraphics[width=0.135\textwidth]{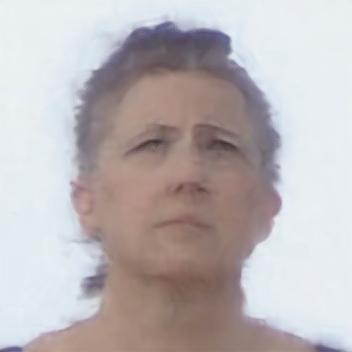}
		\includegraphics[width=0.135\textwidth]{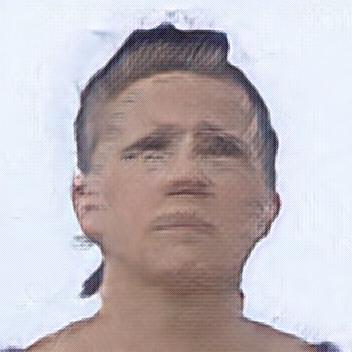}
		\includegraphics[width=0.135\textwidth]{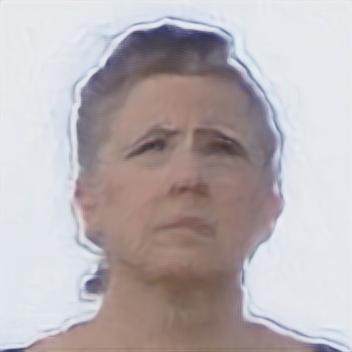}
		\includegraphics[width=0.135\textwidth]{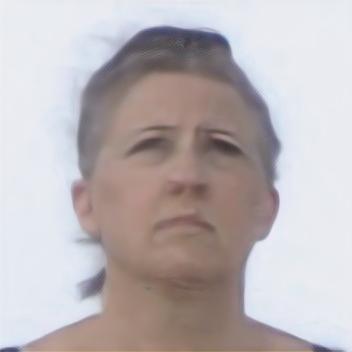}\\ \vskip3pt
		\includegraphics[width=0.135\textwidth]{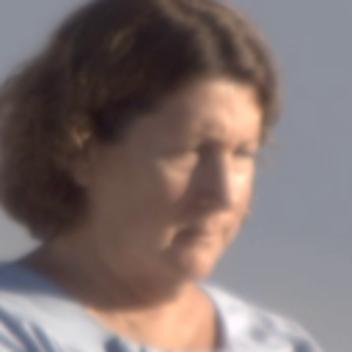}
		\includegraphics[width=0.135\textwidth]{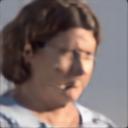} 
		\includegraphics[width=0.135\textwidth]{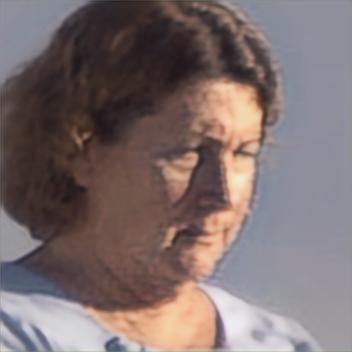}
		\includegraphics[width=0.135\textwidth]{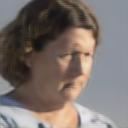}
		\includegraphics[width=0.135\textwidth]{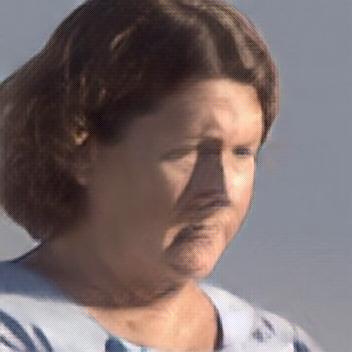}
		\includegraphics[width=0.135\textwidth]{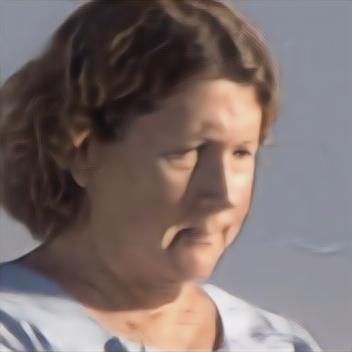} 
		\includegraphics[width=0.135\textwidth]{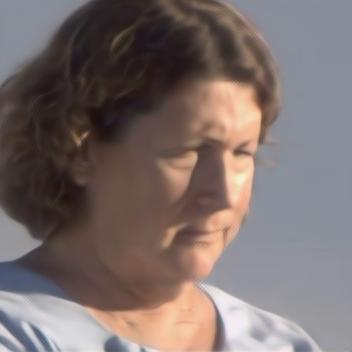}\\ \vskip3pt
		\includegraphics[width=0.135\textwidth]{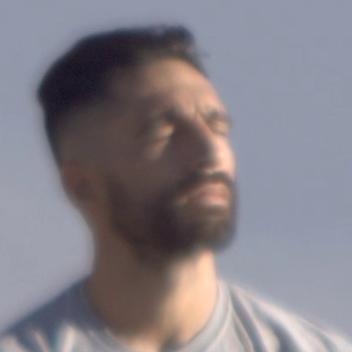}
		\includegraphics[width=0.135\textwidth]{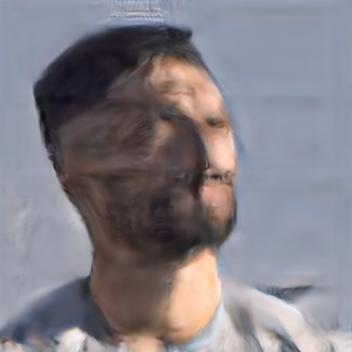} 
		\includegraphics[width=0.135\textwidth]{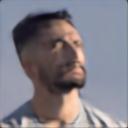}
		\includegraphics[width=0.135\textwidth]{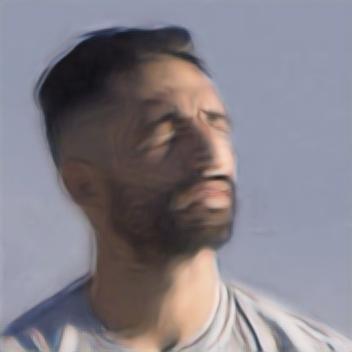}
		\includegraphics[width=0.135\textwidth]{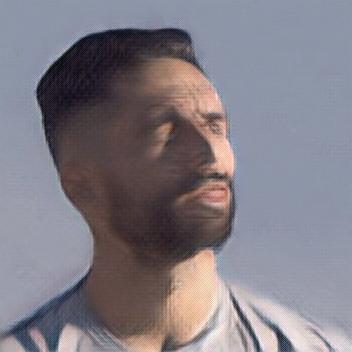}
		\includegraphics[width=0.135\textwidth]{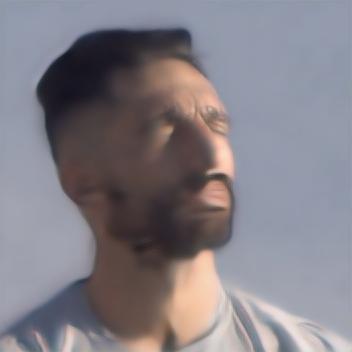} 
		\includegraphics[width=0.135\textwidth]{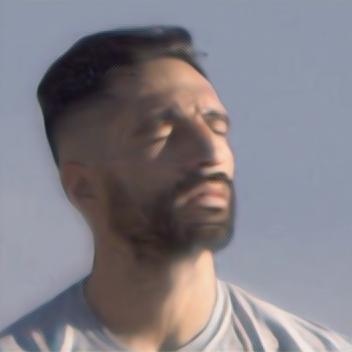}\\
		(a)\hskip60pt(b)\hskip65pt(c)\hskip60pt(d)\hskip60pt(e)\hskip65pt(f)\hskip60pt(g)
		
		\caption{Image restoration results on sample real-world turbulence distorted images. (a) Real turbulence distorted images.  results corresponding to (b) Pan et al.~\cite{pan2014deblurring}(ECCV 2014), (c) Shen et al.~\cite{shen2018deep}(CVPR 2018), (d) Yasarla et al.~\cite{yasarla2019deblurring}(TIP 2020), (e)  Kupyn et al.~\cite{kupyn2018deblurgan}(CVPR 2018), (f) Zhang et al.~\cite{zhang2019deep}(CVPR 2019), (g) Our proposed method TDRN. The method proposed by Pan et al.~\cite{pan2014deblurring} uses face exmplar masks as prior information and produces artifacts on the restored face images.  Shen et al.~\cite{shen2018deep}, and  Yasarla et al.~\cite{yasarla2019deblurring} use face semantic maps as prior information are not able to remove blur and geometric distortions properly.  On the other hand  Kupyn et al.~\cite{kupyn2018deblurgan}, and Zhang et al.~\cite{zhang2019deep} that do not use any prior information produce blurry restored face images. Compared to the other methods, the proposed TDRN method produces sharp and clean face images.}
		\label{Fig:quali_cmp}
	\end{center}
\end{figure*}

\subsection{Results on Real-World Images}	
We also evaluate the performance of different methods on  several real-world turbulence distorted images collected by the US Army, as described in Section~\ref{Train}.  Figure~\ref{Fig:quali_cmp} illustrates the qualitative performance of different methods on two sample real-world turbulence distorted face images from this dataset.  As can  be  seen  from  this  figure,  state-of-the-art restoration methods produce artifacts and blurry outputs especially around the mouth, eyes, and nose regions of the face. On the other hand, TDRN is able to recover details of the face better and significantly improves the visual quality.  


\subsection{Face Recognition}
In order to show the significance of different image reconstruction methods, we perform face recognition on the turbulence restored images. We conduct experiments using both real and synthetic images.  Given a turbulence distorted image, we first restore the image and then perform face recognition using ArcFace~\cite{deng2019arcface}.  The most similar faces (i.e. Top-K nearest matches) for this restored image are selected from the gallery set to check whether they belong to same identity or not.

\noindent {\bf{Results on a synthetic dataset:}}  We use the CelebA dataset for conducting face recognition experiments.
As described earlier, the test set consists of 24,000 images corresponding to 100 different individuals in the CelebA dataset.  Another set of gallery images is created by using 10 different poses from the same 100 identities.  Note that both training and test sets do not contain overlapping images from the same subject.  Face recognition experiment is conducted using these test and gallery sets and the corresponding results are shown in Table~\ref{Table:Facereco}.  As can be seen from this table  that  the restored images by our method have better recognition accuracies than the other state-of-the-art methods. This experiment clearly shows that our method is able to retain the important parts of a face while restoring the image. This in turn helps in  achieving  better  face  recognition  compared  to  the  other methods.

\begin{table}[htp!]
\caption{Top-1, Top-3 and Top-5 face recognition accuracies on the CelebA dataset.} 	
\label{Table:Facereco}
\centering
\resizebox{\textwidth}{!}{
\begin{tabular}{|l|c|c|c|}
	\hline
	\begin{tabular}[c]{@{}c@{}}Deturbulence\\ Method\end{tabular} & Top-1                 & Top-3                 & Top-5                 \\ \hline
	Turbulence-distorted         & 53.05   &    62.38   &  67.90   \\ \hline
	\multicolumn{1}{|l|}{Pix2Pix~\cite{isola2017image}(CVPR 2017)}       &  60.84      &  67.28     &  70.75   \\ \hline
	Pan et al.~\cite{pan2014deblurring}(ECCV 2014)       &  50.88   &   59.61       &    61.39     \\ \hline
	Shen et al.~\cite{shen2018deep}(CVPR 2018)       &  70.11  &   78.42       &  80.77   \\ \hline
	Yasarla et al.~\cite{yasarla2019deblurring}(TIP 2020)       &  78.99     &   84.56    &  88.68    \\ \hline
	Kupyn et al.~\cite{kupyn2018deblurgan}(CVPR 2018)    & 73.49   &  79.50   &  82.44   \\ \hline
	Zhang et al.~\cite{zhang2019deep}(CVPR 2019)       &  80.34     & 86.16      &  89.47      \\ \hline
	\multicolumn{1}{|l|}{TDRN (our method)}       &  \textbf{85.86}      &  \textbf{92.23}     &  \textbf{94.06}    \\ \hline
\end{tabular}}
\end{table}
\begin{table}[htp!]
\caption{Top-1, Top-3 and Top-5 face recognition accuracies on a real-world dataset.} \label{Table:Real_comp}
\centering
\resizebox{\textwidth}{!}{
	\begin{tabular}{|l|c|c|c|}
		\hline
		\begin{tabular}[c]{@{}c@{}} Method\end{tabular} & Top-1                 & Top-3                 & Top-5                 \\ \hline
		Turbulence-distorted         &  38.10  &   49.32   &  56.13   \\ \hline
		\multicolumn{1}{|l|}{Pix2Pix~\cite{isola2017image}(CVPR 2017)}       &   37.82     &   50.76    &  57.41    \\ \hline
		Pan et al.~\cite{pan2014deblurring} (ECCV 2014)      & 35.67    & 45.18         & 50.79        \\ \hline
		Shen et al.~\cite{shen2018deep}(CVPR 2018)       & 39.91   &    52.21      &   58.17  \\ \hline
		Yasarla et al.~\cite{yasarla2019deblurring}(TIP 2020)       &  42.31      &   57.72    &   64.40   \\ \hline
		Kupyn et al.~\cite{kupyn2018deblurgan}(CVPR 2018)    & 40.72   &  54.86   &   62.28   \\ \hline
		Zhang et al.~\cite{zhang2019deep}(CVPR 2019)       &  44.76     &   60.96    &    69.75    \\ \hline
		\multicolumn{1}{|l|}{TDRN (our method)}       &   \textbf{48.73}     &   \textbf{64.41}    &  \textbf{74.32}    \\ \hline
	\end{tabular}}
\end{table}

\noindent {\bf{Results on real-world dataset:}}
The real-world dataset used to conduct experiments in this section is described in subsection~\ref{Train}.  Results corresponding to this experiment are shown in Table~\ref{Table:Real_comp}.  As can be seen from this table, TDRN is able to restore real-world images better and preserves the identities of the subjects in the distorted images better than the other methods.   Note that results on this dataset are worse than the results obtained in the CelebA dataset.  This is mainly due to the fact that, the real dataset contains extreme blur and geometric deformations which makes it a very difficult dataset.  In general, our method is able to achieve $\tilde 4\%$ improvement over the other methods.     

\begin{table}[htp!]
	\caption{PSNR, SSIM, and $d_{VGG}$ results corresponding to the ablation study.} \label{Table:Ablation}
	\centering 
	\resizebox{\textwidth}{!}{
		\begin{tabular}{|l|c|c|c|c|c|c|}
			\hline
			\multirow{2}{*}{\begin{tabular}[c]{@{}c@{}}Deturbulence\\ Method\end{tabular}} & \multicolumn{3}{c|}{CelebA}                                           & \multicolumn{3}{c|}{Helen}                                            \\ \cline{2-7} 
			& PSNR                  & SSIM                  & $d_{VGG}$                & PSNR                  & SSIM                  & $d_{VGG}$             \\ \hline
			Turbulence-distorted   &  22.43    &    0.731   &   5.13      &   22.35     &    0.667    &    6.11     \\ \hline
			Base Network (BN)   &  22.67      &  0.745    &  5.06      &   22.54      &   0.674    &    5.94  \\ \hline
			+ $b$ blur-prior      &  24.01   &  0.781   &   3.84     &   23.91     &  0.714      &    4.42    \\ \hline
			+ both priors ($b$ and $d$)      &    25.04   &    0.802      &   3.57     &    24.47      &   0.725     &    4.17    \\ \hline
			\multicolumn{1}{|l|}{TDRN w/ $\mathcal{L}_{final}$}   &      \textbf{25.42} &    \textbf{0.815}  &   \textbf{3.09}   &    \textbf{25.08} &  \textbf{0.752}       &   \textbf{3.80}    \\ \hline
	\end{tabular}}
\end{table}

\subsection{Ablation Study}

\begin{figure*}[htp!]
	\centering
	\includegraphics[width=0.16\textwidth]{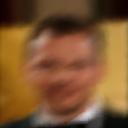}
	\includegraphics[width=0.16\textwidth]{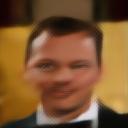} 
	\includegraphics[width=0.16\textwidth]{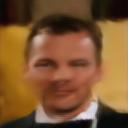}
	\includegraphics[width=0.16\textwidth]{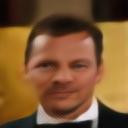}
	\includegraphics[width=0.16\textwidth]{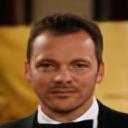}
	\includegraphics[width=0.16\textwidth]{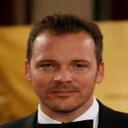}\\ \vskip3pt
	\includegraphics[width=0.16\textwidth]{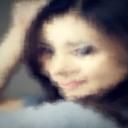}
	\includegraphics[width=0.16\textwidth]{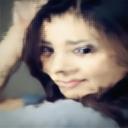} 
	\includegraphics[width=0.16\textwidth]{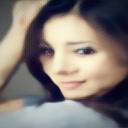}
	\includegraphics[width=0.16\textwidth]{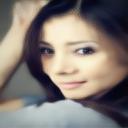}
	\includegraphics[width=0.16\textwidth]{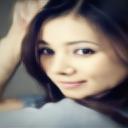}
	\includegraphics[width=0.16\textwidth]{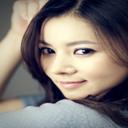}\\
	(a)\hskip75pt(b)\hskip75pt(c)\hskip75pt(d)\hskip75pt(e) \hskip75pt (f)
	\caption{Results of ablation study on two synthetic images.  (a) Real turbulence distorted images.  Results corresponding to (b) Base Netowrk (BN), (c) BN + $b$ blur prior, (d) BN + blur prior $b$  + distortion prior $d$, (e) TDRN w/ $\mathcal{L}_{final}$, and (f) Ground truth clean images.}
	\label{Fig:abl}
\end{figure*}

In order to demonstrate the improvements obtained by different priors and  $\mathcal{L}_{final}$ introduced in the proposed network, we perform an ablation study using the CelebA and Helen datasets involving the following experiments. We start with our base network (BN) which is a combination of UNet~\cite{ronneberger2015u} and DenseNet~\cite{huang2017densely} architectures and then compute the blur prior, $b$ using DBN and pass it with along with the turbulence distorted image. Then we also compute the distortion prior, $d$, using GDRN and pass it along with the turbulence distorted image and $b$. Finally, we train the resultant network with $\mathcal{L}_{final}$ to obtain the proposed network, TDRN. Note that, BN, BN+$b$ (blur prior), BN + both priors ($b$ and $d$) are trained using the L1-loss and the perceptual loss using the turbulence distorted training images.  TDRN is trained using  $\mathcal{L}_{final}$.  The  corresponding  results  are  shown in Table~\ref{Table:Ablation}. Adding the blur prior $b$ as an input to the base network improves the performance by approximately 1.4dB in PSNR. Using the combination of both the distortion prior $d$  and the blur prior $b$ further improved the performance of the base network by 0.8dB.  Finally, when we train the entire network using $\mathcal{L}_{final}$, the performance is further improved by 0.5dB.  Table~\ref{Table:Ablation} clearly shows the importance of different components in our framework.  Sample reconstructions corresponding to the ablation study are shown in  Fig.~\ref{Fig:abl}.  This figure illustrates that in general the quality of image gets better as more components are added to the base network. The best performance is achieved when both priors are used along with $\mathcal{L}_{final}$ to train the network.\\

\noindent {\bf{Experiments regarding blur  and distortion priors:}}  In order to demonstrate the generalizability of our proposed method, i.e computing $b$  and $d$ and using them as prior information for removing atmospheric turbulence distortions, we perform experiments on other state-of-the-art deblurring methods using $b$ and $d$ as priors along with input distorted image to perform deturbulence. In other words, we retrained state-of-the-art deblurring methods with $b$ (blur prior) and $d$ (distortion prior)  as inputs along with turbulence distorted image. As shown in the Table~\ref{Table:exp1}, using our proposed method, i.e using $b$ and $d$ as priors along with input distorted image, the performance of state-of-the-art the methods improves by ~1dB.  The proposed method can be applied to any image restoration network to further improve the performance of the base network.\\ 

\begin{table}[htp!]
	\caption{PSNR comparison on the CelebA test dataset showing the benefits of  $b$ and $d$  on other methods.} \label{Table:exp1}
	\centering
	\resizebox{\textwidth}{!}{
		\begin{tabular}{|l|c|c|c|c|}
			\hline
			\multirow{2}{*}{\begin{tabular}[c]{@{}l@{}}Deturbulence\\ Method\end{tabular}} & \multicolumn{2}{c|}{without $b$ and $d$} & \multicolumn{2}{c|}{with $b$ and $d$} \\ \cline{2-5} 
			& PSNR & SSIM & PSNR & SSIM \\ \hline
			Pix2Pix~\cite{isola2017image}(CVPR 2017) & 22.51 & 0.738 & 23.87 & 0.765 \\ \hline
			Shen et al.~\cite{shen2018deep}(CVPR 2018) & 23.08 & 0.745 & 24.16 & 0.771 \\ \hline
			Kupyn et al.~\cite{kupyn2018deblurgan}(CVPR 2018) & 23.54 & 0.748 & 24.38 & 0.778 \\ \hline
			\multicolumn{1}{|l|}{TDRN (our method)} & 22.67 & 0.745 & 25.04 & 0.802 \\ \hline
		\end{tabular}}
\end{table}

\noindent {\bf{Experiments regarding the loss function  $\mathcal{L}_g$:}} In order to show the benefits of $\mathcal{L}_g$ in training the network, we perform experiments using different combinations of $\{C_i\}_{i=h,v,s}$ in $\mathcal{L}_g$ on turbulence distorted images generated using Helen dataset. In Table~\ref{Table:exp_lg} we trained base network (BN) and TDRN using  different combinations of  $\{C_i\}_{i=h,v,s}$ in $\mathcal{L}_g$, which clearly shows the improvement in performance when we use first and second order gradients with help of corresponding confidence scores.

\begin{table}[htp!]
	\caption{PSNR comparisons using different combinations of $\{C_i\}_{i=h,v,s}$ in $\mathcal{L}_g$ on Helen dataset.} 
	\label{Table:exp_lg}
	\centering
	\resizebox{\textwidth}{!}{
	\begin{tabular}{|l|c|c|c|c|c|}
		\hline
		Network & \multicolumn{1}{l|}{without $\mathcal{L}_g$} & \multicolumn{1}{l|}{$\mathcal{L}_g$ with $C_h$} & \multicolumn{1}{l|}{$\mathcal{L}_g$ with $C_v$} & \multicolumn{1}{l|}{$\mathcal{L}_g$ with $C_s$} & \multicolumn{1}{l|}{$\mathcal{L}_g$ with $\{C_i\}_{i=h,v,s}$} \\ \hline
		BaseNetwork(BN) & 22.35 & 22.67 & 22.76 & 22.62 & 23.11 \\ \hline
		TDRN & 24.47 & 24.78 & 24.83 & 24.65 & 25.08 \\ \hline
	\end{tabular}}
\end{table}

\section{Conclusion}
We proposed a novel method, called TDRN, to address the atmospheric turbulence distortion removal problem from a single image where we exploited blur and distortion priors to obtain better results.  We also proposed a novel loss function, $\mathcal{L}_{final}$, that makes use of the first and the second order image gradients to produce sharper images.  Experiments using synthetic and real data illustrate that this framework is capable of alleviating geometric deformation and blur introduced by turbulence, significantly improving the visual quality of the restored face images.

\ifCLASSOPTIONcompsoc
  \section*{Acknowledgments}
\else
  \section*{Acknowledgment}
\fi

This  research  is  based  upon  work  supported  by  the  Of-ficeof  the  Director  of  National  Intelligence  (ODNI),  Intel-ligenceAdvanced  Research  Projects  Activity  (IARPA),  viaIARPA  R\&D  Contract  No.  2019-19022600002.  The  views and  conclu-sions  contained  herein  are  those  of  the  authors and should not be interpreted as necessarily representing the official  policies or  endorsements,  either  expressed  or  implied, of  the  ODNI, IARPA,  or  the  U.S.  Government.  The  U.S. Government is authorized to reproduce and distribute reprints for Governmental purposes notwithstanding any copyright annotation thereon.

\ifCLASSOPTIONcaptionsoff
  \newpage
\fi

{
	\bibliographystyle{IEEEtran}
	\bibliography{egbib}
}

%
%
%




\end{document}